\documentclass[twoside]{article}

\usepackage[accepted]{claiunconf2023}

\usepackage{amsmath}
\usepackage{amssymb}
\usepackage{algorithm}
\usepackage{algorithmicx}
\usepackage{algpseudocode}
\usepackage{adjustbox}
\usepackage{booktabs}
\usepackage{multirow}
\usepackage{subcaption}
\usepackage{caption}
\usepackage{siunitx}
\usepackage[table]{xcolor}

\definecolor{myrowcolor}{RGB}{255,245,240}  % Modifica i valori RGB come preferisci

\begin{document}

\twocolumn[

\CLAIUnconftitle{Adaptive Hyperparameter Optimization for Continual Learning Scenarios}

\CLAIUnconfauthor{ Rudy Semola, Julio Hurtado, Vincenzo Lomonaco, Davide Bacciu }

\CLAIUnconfaddress{ Department of Computer Science, University of Pisa } ]

%%%%%%%%%%%%%%%%%%%%%%%%%%%%%%%%%%%%%%%%%%%%%%%%%%%%%%%%%%%%%%%%%%%%%%%%%%%%%%%%%%%%%%%%%%%%%%%%%%%%%%%%%%%%%%%%%%%%%%%%%%%%%%%
%
%%%%%%%%%%%%%%%%%%%%%%%%%%%%%%%%%%%%%%%%%%%%%%%%%%%%%%%%%%%%%%%%%%%%%%%%%%%%%%%%%%%%%%%%%%%%%%%%%%%%%%%%%%%%%%%%%%%%%%%%%%%%%%%
\begin{abstract}
Hyperparameter selection in continual learning scenarios is a challenging and underexplored aspect, especially in practical non-stationary environments. Traditional approaches, such as grid searches with held-out validation data from all tasks, are unrealistic for building accurate lifelong learning systems.
This paper aims to explore the role of hyperparameter selection in continual learning and the necessity of continually and automatically tuning them according to the complexity of the task at hand.
Hence, we propose leveraging the nature of sequence task learning to improve Hyperparameter Optimization efficiency. By using the functional analysis of variance-based techniques, we identify the most crucial hyperparameters that have an impact on performance. We demonstrate empirically that this approach, agnostic to continual scenarios and strategies, allows us to speed up hyperparameters optimization continually across tasks and exhibit robustness even in the face of varying sequential task orders. 
We believe that our findings can contribute to the advancement of continual learning methodologies towards more efficient, robust and adaptable models for real-world applications.
\end{abstract}

%%%%%%%%%%%%%%%%%%%%%%%%%%%%%%%%%%%%%%%%%%%%%%%%%%%%%%%%%%%%%%%%%%%%%%%%%%%%%%%%%%%%%%%%%%%%%%%%%%%%%%%%%%%%%%%%%%%%%%%%%%%%%%%%%%%%%%%%%%%%%%%%%%%%%%%%%%%%%%
% 
%%%%%%%%%%%%%%%%%%%%%%%%%%%%%%%%%%%%%%%%%%%%%%%%%%%%%%%%%%%%%%%%%%%%%%%%%%%%%%%%%%%%%%%%%%%%%%%%%%%%%%%%%%%%%%%%%%%%%%%%%%%%%%%%%%%%%%%%%%%%%%%%%%%%%%%%%%%%%%
\section{Introduction}

%%% Context
Hyperparameters play a critical role in the success of machine learning algorithms, allowing practitioners to train or fine-tune models and optimize performance \cite{feurer2019hyperparameter}. Unlike model parameters, which are learned from data during the training process, hyperparameters are set before training begins and determine the model's architecture and optimization settings.
Traditional hyperparameter optimization (HPO) solutions are typically designed to operate under the assumption of independent and identically distributed samples obtained all at once. However, it is often difficult to meet these assumptions in real-world environments, especially when the training data arrives incrementally. 

Continual learning (CL) \cite{parisi2019continual}, which facilitates the progressive training of machine learning models on dynamic and non-stationary data streams \cite{lesort2020continual}, holds promise for enabling companies innovative trend towards Continual Learning \cite{semola2022continual, hurtado2023continual}. However, deploying continual learning to real-world applications remains challenging. Crucially, the role of hyperparameters in continual learning significantly impacts model performance, affecting its ability to generalize and adapt to new tasks while preventing catastrophic forgetting. 
Selecting, adapting, and optimizing hyperparameters are vital to retaining relevant information from past tasks while efficiently acquiring new knowledge.
Nevertheless, the proper selection of hyperparameters is poorly investigated for continual strategies. These last are in many cases found via a grid search, using held-out validation data from all tasks \cite{de2021continual} but are unrealistic for building accurate lifelong learning machines \cite{hemati2023class}.

In this research, we explore relatively uncharted territory with limited prior works, delving into unexplored hypotheses that require further validation. Our effort is focused on addressing the following \textbf{research questions}, which we categorize into the dimensions of performance, efficiency, and robustness.
 \textbf{(RQ1)} How important is each of the hyperparameters, and how do their values affect performance in sequence learning? Which hyperparameter interactions matter? How do the values of the hyperparameters differ across different incremental learning steps? How do the answers to these questions depend on the tasks similarity of the sequence under consideration?
 \textbf{(RQ2)} Given that conducting HPO on each individual experience leads to improved performance but comes at the expense of computational resources, how can we enhance efficiency by leveraging hyperparameter importance information?
\textbf{(RQ3)} What is the effect of HPO for each model update in relation to its robustness to the sequence order?

% (2) the aims of the paper ("scientific value": novelty, scientific contribution)
A primary objective of this paper is to delve into the pivotal role of hyperparameters optimization in the context of continual learning and examine possible strategies for their effective management.
To summarize, we expect that \textbf{our main contributions} should be:
\begin{enumerate}
    \item Introducing and defining the problem of optimising hyperparameters through the learning experience with a realistic evaluation protocol, thereby enabling comprehensive study in this field.
    \item To quantitatively assess the significance of individual hyperparameters and their interactions within each task, we propose to use the functional analysis of the variance (\emph{fANOVA}) technique, outlined in \cite{hutter2014efficient, van2018hyperparameter}, for a sequence of learning tasks. The expected result should provide a quantitative basis to focus efforts on automated adaptive hyperparameter optimization for continual learning strategies. 
    \item To present a novel hyperparameter update rule that significantly enhances the rapid adaptation process over streaming data. Our approach introduces an adaptive policy that automatically selects task-conditioned hyperparameters on a per-step basis, thereby greatly improving the optimization effectiveness. 
    \item Through experiments on well-established benchmarks and baseline, we want to explore and compare our results with significant improvements in terms of average accuracy and computational cost (performance-efficiency trade-off).
    \item To examine our proposal solution with existing baseline approaches in terms of robustness to the sequence order. The expected result should provide an improvement of this dimension.
\end{enumerate}

% (3) ToC: Structure of the paper
%TODO Structure of paper

%%%%%%%%%%%%%%%%%%%%%%%%%%%%%%%%%%%%%%%%%%%%%%%%%%%%%%%%%%%%%%%%%%%%%%%%%%%%%%%%%%%%%%%%%%%%%%%%%%%%%%%%%%%%%%%%%%%%%%%%%%%%%%%
%
%%%%%%%%%%%%%%%%%%%%%%%%%%%%%%%%%%%%%%%%%%%%%%%%%%%%%%%%%%%%%%%%%%%%%%%%%%%%%%%%%%%%%%%%%%%%%%%%%%%%%%%%%%%%%%%%%%%%%%%%%%%%%%%
\section{Background}
% (Related Works)

\subsection{Hyperparameter Optimization (HPO)}
Hyperparameter optimization (HPO) is a critical process in machine learning and statistical modelling, aimed at finding the optimal values for hyperparameters that govern the behaviour of a learning algorithm \cite{feurer2019hyperparameter}. 
The goal is to find the hyperparameter configuration that maximizes the model's performance on a validation set or minimizes a chosen objective function.

Given a learning algorithm $\mathcal{A}$ with a set of hyperparameters $\mathcal{H}$ and a performance metric $\mathcal{M}$, the hyperparameter optimization problem aims to find the optimal hyperparameter configuration $\mathbf{h}^* \in \mathcal{H}$ that maximizes the performance metric $\mathcal{M}$ on a validation set. Formally, the problem can be defined as:
\begin{equation}
\mathbf{h}^* = \arg\max_{\mathbf{h} \in \mathcal{H}} \mathcal{M}(\mathcal{A}(\mathcal{D}_{\text{tr}}, \mathbf{h}), \mathcal{D}_{\text{val}})
\label{eq:hpo_math} 
\end{equation}
where $\mathcal{D}_{\text{tr}}$ is the training dataset, $\mathcal{A}(\mathcal{D}_{\text{tr}}, \mathbf{h})$ represents the trained model obtained by applying algorithm $\mathcal{A}$ on the training dataset with hyperparameter configuration $\mathbf{h}$, and $\mathcal{M}(\cdot)$ measures the performance of the model on the validation dataset $\mathcal{D}_{\text{val}}$.

\subsubsection{Hyperparameter Importance}
Past knowledge about hyperparameter importance is mainly based on a combination of intuition, own experience and folklore
knowledge. However, a more data-driven and quantitative approach has been introduced through the functional ANOVA framework (\emph{fANOVA}), as presented in the work by Hutter et al. \cite{hutter2014efficient}. This framework enables the analysis of hyperparameter importance, offering valuable insights into the factors influencing performance.

The fundamental idea underlying hyperparameter importance lies in the extent to which a particular hyperparameter contributes to the variance in performance. Hyperparameters that significantly affect performance outcomes require careful tuning to achieve optimal results. Conversely, hyperparameters with minimal impact are considered less crucial and may warrant less attention during tuning.
Functional ANOVA not only attributes variance to individual hyperparameters but also explores the interaction effects among sets of hyperparameters. This analysis sheds light on which hyperparameters can be tuned independently and which exhibit interdependence, necessitating joint tuning for optimal performance.

The study by Hutter et al. \cite{hutter2014efficient} applies functional ANOVA to the outcomes of a single hyperparameter optimization procedure conducted on a single dataset. Building upon this framework, Van Rijn et al. \cite{van2018hyperparameter} extended the concept of hyperparameter importance to assess the general importance of hyperparameters across multiple datasets, providing broader insights into their significance.

\subsection{Continual Learning (CL)}
Continual Learning strategies are efficient incremental training strategies \cite{semola2022continual, de2021continual}. 
An alternative approach would be to fine-tune the previous model on the new sequence of data. 
However, this could lead to so-called catastrophic forgetting, where performance on older data deteriorates while training on new data. 
The incremental learner has two goals to address this issue: to effectively learn the current task (plasticity) while retaining performance on all previous tasks (stability).
Continual learning solutions have to strike a compromise between these two extremes.

Following the definitions in \cite{de2021continual}, the goal of Continual Learning is to train a model over a \--- possibly infinite \--- sequence of tasks or experiences. Each task can be represented as dataset $ D(x^{(t)},y^{(t)}) $ sample from a distribution $ D^{(t)} $ in time $t$, were $x^{(t)}$ is the set of samples for task $t$, and $y^{(t)}$ the corresponding label. 
As a result, the learner function should be able to model the cumulative probability distribution of the data $\mathcal{P}(Y|X)$ where $X$ is the set of all samples for all tasks and $Y$, are their corresponding labels. Since this probability distribution is intractable, continual strategies perform indirect optimization by minimizing the following: 

\begin{equation}
    \sum_{t=1}^{\mathcal{T}} \mathbb{E}_{x^{(t)},y^{(t)}} [ \mathcal{L}(\mathcal{A}_t^{CL}(x^{(t)}; \theta), y^{(t)})]
\end{equation}

\noindent with limit or no access to previous data $ (x^{(t')},y^{(t')}) $ when training tasks $t > t'$. Then, Continual Learning methods seek to optimize the parameters $\theta$ by minimizing the loss expectancy for all tasks in the sequence $ \mathcal{T} $.

\subsection{HPO in Continual Learning Scenarios}

\subsubsection{Problem Formulation}
In continual learning, models are trained and applied for prediction without having all training data beforehand.
The process that generates the data may change over time, leading to concept drift, a usually unpredictable shift over time in the underlying distribution of the data.
It is crucial to understand the dynamics of concept drift and its effect on the search strategy used by the HPO technique in order to design a successful strategy in a non-stationary sequence data stream.
Considering the equation \ref{eq:hpo_math} designed for stationary environments, HPO for a particular task $t$ in sequence learning $\mathcal{T} = t_0,..., t_M = \{ t\}$ can be described as the following optimization problem:
\begin{equation}
\mathbf{h}_t^* = \arg\max_{\mathbf{h} \in \mathcal{H}} \mathcal{M}(\mathcal{A}_t^{CL}(\mathcal{D}^{(t)}_{\text{tr}}, \mathbf{h}), \mathcal{D}_{\text{val}}) 
\end{equation}
Here $\mathcal{D}_{\text{val}}$ is the subset of the validation set for the current task and previous tasks and $\mathcal{A}_t^{CL}$ is the continual learner achieved so far with hyperparameter configuration $\mathbf{h}$ sought only on currently available dataset $\mathcal{D}^{(t)}_{\text{tr}}$.

\subsubsection{Related Works}

Hyperparameters play a significant role in continual learning by influencing the model's ability to adapt to new tasks while retaining knowledge from previous tasks.
Note that strategies tackling the continual learning problem typically involve extra hyperparameters to balance the stability plasticity trade-off.
These hyperparameters are in many cases found via a grid search, using held-out validation data from all tasks \cite{de2021continual}. 
In particular, the typical assumption is to have access to the entire stream at the end of training for model selection purposes.
After that, as a common practice, the hyperparameters are set to a fixed value throughout all incremental learning sessions.
One simple motivation is that tuning hyperparameters is a major burden and is no different in the continual setting. Another motivation is the simplicity of implementation. 
Nevertheless, this inherently violates the main assumption in continual learning, namely no access to previous task data and in practical scenarios is unrealistic for building accurate lifelong learning machines \cite{hemati2023class, chaudhry2018efficient}.
In the following, we present a comprehensive summary of related works concerning adaptive HPO and CL, categorizing them into three classes.

\textbf{Dynamic Task-Specific Adaptation.} 
Real-world learning presents a dynamic environment where the optimal hyperparameter configuration may evolve over time. In response to this challenge, recent work by Gok et al. \cite{gok2023adaptive} focuses on continual learning and investigates the necessity of adaptive regularization in Class-Incremental Learning. This approach dynamically adjusts the regularization strength based on the specific learning task, avoiding the unrealistic assumption of a fixed regularization strength throughout the learning process. Empirical evidence from their experiments highlights the significance of adaptive regularization in achieving enhanced performance in visual incremental learning.
Concurrently, Wistuba et al. \cite{wistuba2023renate} address the challenge of practical HPO for continual learning. They propose adjusting hyperparameters such as learning rates, regularization strengths, or architectural choices for different tasks, allowing the model to adapt its behaviour to each task's specific requirements. Their empirical findings demonstrate improvements in performance through this approach.
However, it is worth noting that despite the demonstrated performance gains, Wistuba et al. \cite{wistuba2023renate} do not fully leverage the inherent nature of continual learning problems or exploit the potential knowledge transfer from previous HPOs. Incorporating transfer learning techniques and capitalizing on insights from prior tasks could further enhance the model's performance and overall efficiency in continual learning scenarios.

\textbf{Transfer Learning and Knowledge Distillation.} \cite{stoll2020hyperparameter} has explored the idea of adapting hyperparameters to optimize performance for individual tasks, facilitating automatic knowledge transfer from previous HPO endeavours across datasets. Furthermore, hyperparameters and dynamical architecture chance can facilitate transfer learning, where knowledge and hyperparameters learned from previous tasks are leveraged to improve performance on new tasks. Knowledge distillation techniques can be applied \cite{rusu2016progressive, rosasco2021distilled}, where hyperparameters guide the transfer of knowledge from a larger or more accurate model to a smaller or more specialized model.

\textbf{Empirical Studies.} Hyperparameters can govern the use of memory replay or experience replay mechanisms to mitigate catastrophic forgetting. The hyperparameters determine aspects such as the importance of past data samples, the frequency of replay, or the balance between old and new data, thereby influencing the impact of replay on model performance as highlighted in \cite{merlin2022practical, de2021continual, hemati2023class}.

%%%%%%%%%%%%%%%%%%%%%%%%%%%%%%%%%%%%%%%%%%%%%%%%%%%%%%%%%%%%%%%%%%%%%%%%%%%%%%%%%%%%%%%%%%%%%%%%%%%%%%%%%%%%%%%%%%%%%%%%%%%%%%%
%
%%%%%%%%%%%%%%%%%%%%%%%%%%%%%%%%%%%%%%%%%%%%%%%%%%%%%%%%%%%%%%%%%%%%%%%%%%%%%%%%%%%%%%%%%%%%%%%%%%%%%%%%%%%%%%%%%%%%%%%%%%%%%%%
\section{Adaptive Hyperparameter Optimization for Continual Learning Scenarios}

\subsection{Methodology and Key Assumptions}
%Methodology with the assumptions behind (motivation)
We hypothesize that the assumption of hyperparameters constancy in all sequence stream data is unrobust and ineffective in continual learning settings.
Firstly, the common practice of using held-out validation data from all tasks inherently violates the main assumption of having no access to previous task data.
Secondly, the assumption of fixed hyperparameters in all sequence stream data is unrealistic for building effective continual learning real-world systems.
Finally, assuming to work on more practical scenarios, optimizing hyperparameters over the entire data sequence is not possible and real solutions involve changing or not the hyperparameter when it detects a distribution shift or model's performance decay. 

\begin{algorithm}
    \caption{Adaptive Hyperparameters Tuning for Continual Learning}
    \begin{algorithmic}[1]
        \Require{$\mathcal{H}=\{H_n\}$ configuration space with hyperparameters $N$; $\mathcal{T}=\{t\}$ sequence tasks; $\mathcal{A}^{CL}$ continual learner}
        \State \textbf{Initialization}
        \For{$(t,i)$ \textbf{in} $\mathcal{T}$}
            \If{$i < m$}
                \Comment{First \emph{m} tasks}
                \State $h_t^* = \text{hpo}(\mathcal{H} , t, \mathcal{A}_t^{CL})$
                \State $\{\mathcal{H}_n, \mathrm{v}\} = \text{get\_param\_imp}(\text{\emph{f}ANOVA}, \text{hpo}, \mathcal{H})$
                \State $\mathcal{H}^k = \text{top\_k\_hp}(\{\mathcal{H}_n, \mathrm{v}\}, k)$
            \Else
                \Comment{Rest of other tasks}
                \State $h_t^* = \text{hpo\_warm\_start}(\mathcal{H}^k, h_{t-1}^*, t, \mathcal{A}_t^{CL})$
            \EndIf
        \EndFor
        \State \textbf{return} $h_\mathcal{T}^* \text{ best configuration in all the sequence tasks}$
    \end{algorithmic}
    \label{alg:hpo_math} 
\end{algorithm}

% TODO Methodology with the assumptions behind (motivation): MORE DETAIL OF METHODOLOGIES
The idea behind Adaptive Hyperparameters Tuning for Continual Learning  (Algorithm \ref{alg:hpo_math}) is simple.
In the first \emph{m} tasks, we select the best configuration in $\mathcal{H}$ as in a typically stationary setting.
Exploiting the \emph{fANOVA} evaluator, we compute parameter importances based on completed trials in the given HPO and associated continual learner $\mathcal{A}_t^{CL}$.
In the remaining tasks, we speed up the optimization process based on the importance of each parameter. 
The proposal method automatically selects the $k$ most important parameters to be changed and keeps fixed the others with the optimal value computed in the previous task. 
Note that this policy selection of the parameters to tune is automatic and agnostic to CL strategies and sequence tasks (and their order).

$\emph{hpo}$: Implement a specific HPO for each model update, i.e. grid search, population-based training or Bayesian optimization. The latter should speed up the HPO if we start from the best configuration in the previous task. 
For this reason, we want to use Bayesian Optimization and Hyperband (BOHB) \cite{falkner-icml-18} that performs robust and efficient hyperparameter optimization at scale by combining the speed of Hyperband searches with the guidance and guarantees of convergence of Bayesian Optimization among tasks.
%For this reason, we want to use Tree-structured Parzen Estimator (TPE) \cite{TPE} that performs robust and efficient hyperparameter optimization at scale by combining the speed of Hyperband searches with the guidance and guarantees of convergence of Bayesian Optimization among tasks.

$\emph{get\_param\_imp}$: Evaluate parameter importances based on \emph{fANOVA} evaluator in the given HPO and $\mathcal{A}_t^{CL}$. The function returns the parameter importances as a dictionary $\{\mathcal{H}_n, \mathrm{v}\}$ where the keys consist of specific parameter $H_n$ and values importance $\mathrm{v} \in \{0,1\}$.  

$\emph{top\_k\_hp}$: function that return $\mathcal{H}^k$ as a subspace of $\mathcal{H}$ after ordering the parameters by importance.

$\emph{hpo\_warm\_start}$: the idea is to speed up the tuning process among the sequence tasks by starting from the optimal parameters found in the previous task $ h_{t-1}^*$. Furthermore, knowing the importance of each parameter we select the subspace of $\mathcal{H}$ with the most \emph{k} important to be tuned and keep fixed the others. 

%Our idea is to exploit this hyperparameter importance information to automatically select the parameters to be tuned and which could be fixed dynamically in the non-stationary learning sequence.
We believe that the tasks in the sequence, even if they have different distributions, have enough task similarity to be exploited in the sequence HPO to efficiently tune the parameters on the current task. To do this, our idea is to exploit the hyperparameter importance information to automatically select the parameters to be tuned and which could be fixed dynamically in the non-stationary learning sequence.

\subsection{Experimental protocol}
 \label{sec:experimental-protocol} 
The goal of this section is to describe the protocol, benchmarks, baseline, and continual strategies that we plan to use in the experiments. 

\subsubsection{Benchmarks and Metric}
 \label{sec:benchmarks-metric} 
Continual learning algorithms are evaluated by \emph{benchmarks}: they specify how the stream of data is created by defining the originating dataset(s), the number of samples, the criteria to split the data into different tasks and so on. 
In literature, different benchmarks are used to evaluate results.

In this paper, we would conduct experiments using benchmarks from two distinct scenarios: Class Incremental, where each task introduces new classes without revisiting old ones in the training stream, and Domain Incremental, where each task presents new instances for existing classes without reusing old instances in the training stream \cite{lesort2020continual}.
We have selected three for \textbf{Class Incremental Learning}: Split-CIFAR10 \cite{zenke2017continual}, Split-TinyImageNet \cite{mai2022online} and CORe50-NC \cite{lomonaco2017core50}. 
These benchmarks are derived respectively from CIFAR-10 \cite{krizhevsky2009learning} and TinyImagenet \cite{le2015tiny} datasets while CORe50-NC is a benchmark specifically designed for continual learning.
For \textbf{Domain Incremental Learning} we have selected Rotated-MINIST derived from the MNIST dataset \cite{deng2012mnist} and CORe50-NI.
Each dataset exhibits visual classification learning.
We will train all the models with Split-CIFAR10 and Rotated-MINIST online, while the others will be in batch mode.
We will resort to the standard metrics for evaluation, i.e. accuracy, which measures the final performance averaged over all tasks (also defined as \emph{stream accuracy}, SA) \cite{lesort2020continual}.
In both incremental scenarios, higher values indicate better performance.

\subsubsection{Baselines}
We intend to compare our solution with two opposite approaches used both in literature \cite{wistuba2023renate} and more practical scenarios for hyperparameter optimization for sequence task learning.
\begin{itemize}
    \item The upper bound in terms of performance and lower bound for computational cost is HPO for each model update as used in \cite{wistuba2023renate}. We plan to use grid search or Bayesian optimization as an HPO technique to improve as possible the performance comparison. The computational cost expected to achieve is $|\mathcal{H}|* |\mathcal{T}|$ which is the worst case.
    \item The lower bound in terms of performance and upper bound for computational cost is to perform HPO only in the first experience and keep fixed the configuration for the rest of the sequence learning. The computational cost expected to achieve is $|\mathcal{H}|$ which is the best case.
\end{itemize}

\subsubsection{Continual Learning strategies}
We selected five strategies among the most popular and promising rehearsal and regularization approaches.

%rehearsal
In particular, for rehearsal, we intend to use the following.

\textbf{Experience Replay (ER).} We selected Replay \cite{buzzega2021rethinking, merlin2022practical} because it is an effective continual strategy for practical scenarios. 
In particular, it is a simple way to prevent catastrophic forgetting, and it performs better with respect to more complicated strategies. In our future experiments, we plan to explore different buffers with different policies to select-discard samples.

\textbf{Greedy Sampler and Dumb Learner (GDumb)} \cite{prabhu2020gdumb} is a simple approach that is surprisingly effective.
Compared to other rehearsal methods, with the same memory size, this strategy is more efficient, in terms of execution time and resources. In a particular setting, this simple strategy can outperform other approaches.

\textbf{Dark Experience Replay (DER/DER$++$)} \cite{buzzega2020dark}, as a more recent replay method, relies on dark knowledge for distilling past experiences, sampled over the entire training trajectory. With respect to ER, DER converges to flatter minima and achieves better model calibration at the cost of limited memory and training time overhead.

%regularization
For well-established regularization methods, we intend to employ the following category of continual learners because we believe they are more sensitive to hyperparameter selection and necessitate hyperparameters that adapt dynamically throughout the learning sequence.

\textbf{Learning-without-Forgetting (LwF)} \cite{li2017learning} is a knowledge-distillation approach where the teacher branch is the model from the previous task, and the student branch is the current model. 
The aim is to match the activations of the teacher and student branches, either at the feature or logit layer.

\textbf{Synaptic Intelligence (SI).} \cite{zenke2017continual} introduces intelligent synapses that bring some of this biological complexity into artificial neural networks. Each synapse accumulates task-relevant information over time and exploits this information to rapidly store new memories without forgetting old ones. 

\begin{table}[hbt!]
\caption{The continual strategies with their specific and general hyperparameters. $mem\_size$: replay buffer size; (*) both the optimizer and model could have further parameters.}
\centering
    \begin{adjustbox}{width=.48\textwidth}
    \small
        \begin{tabular}{ll}
        \toprule
            \textbf{Strategy} & \textbf{Hyperparameters} \\
        \midrule
            ER (replay) & $mem\_size$ \\ &  \emph{buffer type} \\ & \emph{storage policy} \\
        \midrule
            GDumb &  $mem\_size$ \\
        \midrule
            DER/DER$++$ &  $mem\_size$ \\ & \emph{alpha} \\ & $beta=0$ (DER) \\ & \emph{with beta} $\neq$ 0 (DER++) \\
        \midrule
            LwF &  \emph{alpha} \\ & \emph{temperature} \\
        \midrule
            SI & \emph{lambda} \\ & \emph{eps} \\
        \midrule
            General Hyperparameters &  \emph{optimizer}* \\
            &  \emph{lr} \\
            &  \emph{training epochs} \\
            &  \emph{batch size} \\
            &  \emph{model}* \\
        \bottomrule
        \end{tabular}
    \end{adjustbox}
\label{tab:cl-hps}
\end{table}

We report in Table \ref{tab:cl-hps} the selected ad-priori most interesting hyperparameters for each continual method and additional hyperparameters less specific to the learner. This selection for the experiments has a twofold intention. Primarily, we aim to spotlight parameters specific to individual continual learners, particularly pertinent to addressing RQ1. Additionally, we recognize the significance of general parameters like learning rate, which should impact overall performance. Therefore, we intend to showcase their inclusion as a demonstration of the adaptability of our solution across a wide array of incremental scenarios and strategy types.

\subsubsection{Implementation details}
We will run experiments on three different seeds and report their average. 
For each benchmark, the evaluation protocol will be split by pattern. First, we will split the overall dataset into 90\% model selection and 10\% model assessment patterns. Then, we will use 15\% of model section data as a validation set and a batch size of 32 examples.
For the experimental part, we intend to use Avalanche \cite{lomonaco2021avalanche} the reference continual learning framework based on PyTorch and Ray Tune for the hyperparameter optimization \cite{liaw2018tune}.
To quantitatively assess the significance of individual hyperparameters and their interactions within each task, we will leverage various metrics, as outlined in \cite{hutter2014efficient, van2018hyperparameter}, and will implement them within the Optuna framework \cite{akiba2019optuna}.

%Finally, for the analysis of the robustness with respect to sequence order, we will run experiments on three different orders and will report their average with standard deviation. 
Finally, for the analysis of the robustness with respect to sequence order, we will conduct experiments across three distinct orders on the benchmarks defined in Subsection \ref{sec:benchmarks-metric} and subsequently report their average with associated standard deviations. We intend to employ the stream accuracy metric for each model update, the outcomes of which will be visually presented in plots that facilitate a comparative analysis of our solution against all baseline methods. 

%%%%%%%%%%%%%%%%%%%%%%%%%%%%%%%%%%%%%%%%%%%%%%%%%%%%%%%%%%%%%%%%%%%%%%%%%%%%%%%%%%%%%%%%%%%%%%%%%%%%%%%%%%%%%%%%%%%%%%%%%%%%%%%
%
%%%%%%%%%%%%%%%%%%%%%%%%%%%%%%%%%%%%%%%%%%%%%%%%%%%%%%%%%%%%%%%%%%%%%%%%%%%%%%%%%%%%%%%%%%%%%%%%%%%%%%%%%%%%%%%%%%%%%%%%%%%%%%%
\section{Expected Results and Discussion}

Our objective is to empirically demonstrate that: 
\begin{itemize}
    \item The hypothesis of automatically adjusting the most effective continuous parameters for each update can significantly expedite the accuracy of the overall sequence task.
    \item The quite task-independent hyperparameters can be computed from the initial tasks in a similar sequence of streamed data and can considerably accelerate the hyperparameter processes in practical scenarios while maintaining an advantageous performance-efficiency tradeoff.
    \item By striking a favourable balance between performance and efficiency, we hypothesize that our proposal method will greatly enhance robustness, particularly in terms of sequence order. 
\end{itemize}

% Idea of showing and expected results (highlight)
\subsection{Important and Adaptive Hyperparameters for CL Analyses}
 \label{subsec:rq1} 

\textbf{(RQ1)} How important is each of the hyperparameters, and how do their values affect performance in sequence learning? Which hyperparameter interactions matter? How do the values of the hyperparameters differ across different incremental learning steps? How do the answers to these questions depend on the task similarity of the sequence under consideration?

We want to demonstrate that performance variability is often largely caused by a few hyperparameters that define a subspace to which we can restrict configuration. Moreover, given a specific continual learner and sequence of similar tasks, we believe that this small set of hyperparameters responsible for the most variation in performance is the same set of hyperparameters across the sequence.
%We want to show here the most interesting results on benchmarks in a table with all the hyperparameters used for each strategy.

% Idea of showing and expected results (highlight)
\subsection{Performance-efficiency Analyses}
 \label{subsec:rq2} 

\textbf{(RQ2)} Given that conducting HPO on each individual experience leads to improved performance but comes at the expense of computational resources, how can we enhance efficiency by leveraging hyperparameter importance information?

According to similar work like \cite{van2018hyperparameter}, we want to demonstrate empirically that the hyperparameters determined as the most important ones indeed are the most important ones to optimize also in sequence task learning. In particular, given a continual learner, only a small set of hyperparameters are responsible for most variation in performance and this is the same for all tasks in a sequence. As a result, we only perform hyperparameter optimization in $\mathcal{H}$ for the initial $m$ tasks with $m < |\mathcal{T}|$ and speed up the hyperparameter optimization through the remaining data stream.
%Here the idea is to show the final accuracy of the different benchmarks and strategies. We can add plots of accuracy among tasks.
%Moreover, we compute the timing required in hpo for each task and report them in plots among the tasks.

% Idea of showing and expected results (highlight)
\subsection{Robustness Analyses}
 \label{subsec:rq3} 

\textbf{(RQ3)} What is the effect of HPO for each model update on robustness to the sequence order?

Based on the insights from \cite{wistuba2023renate}, we intend to conduct further experiments to emphasize the importance of performing HPO for each model update and emphasize its robustness concerning the sequence order.
We anticipate observing enhanced performance trends across the stream and reduced variance.

%%%%%%%%%%%%%%%%%%%%%%%%%%%%%%%%%%%%%%%%%%%%%%%%%%%%%%%%%%%%%%%%%%%%%%%%%%%%%%%%%%%%%%%%%%%%%%%%%%%%%%%%%%%%%%%%%%%%%%%%%%%%%%%
%  
%%%%%%%%%%%%%%%%%%%%%%%%%%%%%%%%%%%%%%%%%%%%%%%%%%%%%%%%%%%%%%%%%%%%%%%%%%%%%%%%%%%%%%%%%%%%%%%%%%%%%%%%%%%%%%%%%%%%%%%%%%%%%%%
\section{Changes and Additions to the pre-registered proposal}

Given the time constraints and our obtained results, we have chosen to conduct a comprehensive analysis of hybrid and traditional continual learning strategies, emphasizing their robustness in per-task performance in conjunction with HPO.
This decision involved focusing on four benchmarks, i.e. Rot-MNIST, Split-CIFAR10, Split-Tiny and CORe50 within a DIL setting.
Moreover, the effort made in exploring the importance of hyperparameters for these four benchmarks has improved the analysis of how to tune the parameters better for hybrid and traditional continual learning strategies. 
Moreover, we have decided to change the optimizer, specifically the Tree-structured Parzen Estimator (TPE) \cite{TPE}, instead of BOHB. The main reason for this change is that TPE is highly efficient in searching for hyperparameters and can find good sets with relatively few evaluations compared to BOHB.
Finally, we have named the two \textit{baselines}. The first baseline, which represents the upper bound in terms of accuracy but comes with a high computational cost, is called \emph{HPO} while the second baseline is called \emph{Fixed}. 

%\begin{itemize}
%    \item \textbf{Fixed.} The first baseline random choice of the hyperparameters in the first task and kept fixed in the sequence. We consider this baseline the lower bound in terms of performance and call it \emph{Fixed} for the rest of the paper. 
%    \item \textbf{HPO.} This second baseline performs the tuning of hyperparameters per task using TPE as an optimization strategy. This baseline is the upper bound in terms of accuracy but with a high computational cost, that we aim to reduce with our approach. We call it \emph{HPO} for the rest of the paper. 
%\end{itemize}

%%%%%%%%%%%%%%%%%%%%%%%%%%%%%%%%%%%%%%%%%%%%%%%%%%%%%%%%%%%%%%%%%%%%%%%%%%%%%%%%%%%%%%%%%%%%%%%%%%%%%%%%%%%%%%%%%%%%%%%%%%%%%%%
%  
%%%%%%%%%%%%%%%%%%%%%%%%%%%%%%%%%%%%%%%%%%%%%%%%%%%%%%%%%%%%%%%%%%%%%%%%%%%%%%%%%%%%%%%%%%%%%%%%%%%%%%%%%%%%%%%%%%%%%%%%%%%%%%%
\section{Results and Discussion}
In this section, we provide the results of the experiments with analysis and discussion. 
We use Rot-MNIST and CIFAR10 for Online Incremental Scenarios (single epoch over the same task) while Tiny and CORe50 for Batch Incremental Scenarios (multiple epochs).

\subsection*{Experimental Setup}
This subsection covers additional information on the experiment setting not covered in the \ref{sec:experimental-protocol}.

\textbf{HW usage.} To guarantee a fair comparison among all the methods, we conduct all tests under the same conditions, running each benchmark on a Multi-GPU NVIDIA-SMI server with 80-core Intel Xeon CPU E5-2698 v4 and 4 Tesla V100 GPUs 11.2 CUDA Version.

\textbf{Architecture.} For the experiments in Rot-MNIST benchmark, we employ a fully connected network with two hidden layers of 126 units each, followed by a ReLU layer.
For CIFAR10, Tiny ImageNet and CORe50, we use Slim-ResNet18 (not pre-trained). Both these models are available in the Avalanche library.
%NOTA: if required ref for ResNet we should see the DER paper (15)...

\textbf{Training and Hyperparameter Optimization.} To provide a fair comparison among all the methods, we train all the networks using the Adam with decoupled weight decay (AdamW) \cite{adamw} optimizer. For Rot-MNIST and CIFAR10 settings, we conduct experiments with one epoch per task (online)  and a buffer memory size of 500. Conversely, we increase the number of epochs to 50 for Tiny and 20 for CORe50 with early stopping and 5120 as buffer memory for the batch incremental scenario. We use TPE as the Hyperparameter optimization method for all the experiments.

%%%%%
\subsection{Important and Adaptive Hyperparameters for Continual Learning Analyses}
This section delves into the discussion of experimental results achieved for \textbf{(RQ1)}, with further discussion on our early hypotheses provided in Subsection \ref{subsec:rq1}.
By studying how important is each of the hyperparameters, and how their values affect performance in sequence learning, we explore how their contribution differs on overall sequence and across different tasks.

\subsubsection{Hyperparameter Importance on Overall Sequence Performance}

%%% IMP AVG  %%%
\begin{figure*}[h!]
  \centering
  
  \begin{subfigure}[b]{0.4\textwidth}
    \captionsetup{justification=centering}
    \caption{\label{fig:imp_er_online} ER and DER, Split-CIFAR10}
    \includegraphics[width=\textwidth]{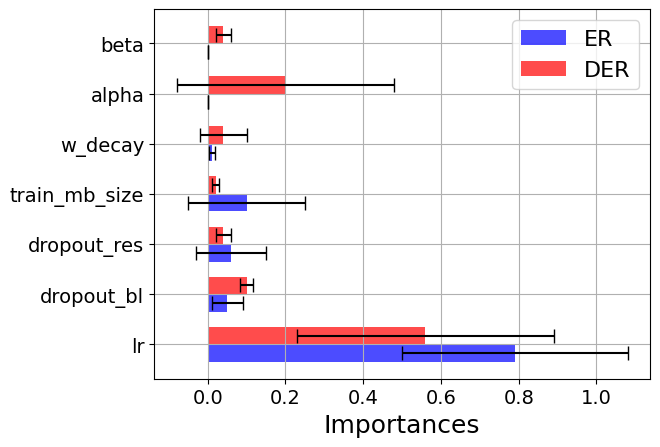}
  \end{subfigure}%
  \hspace{0.75cm}
  \begin{subfigure}[b]{0.4\textwidth}
    \captionsetup{justification=centering}
    \caption{\label{fig:imp_er_online} ER and DER, Rot-MNIST}
    \includegraphics[width=\textwidth]{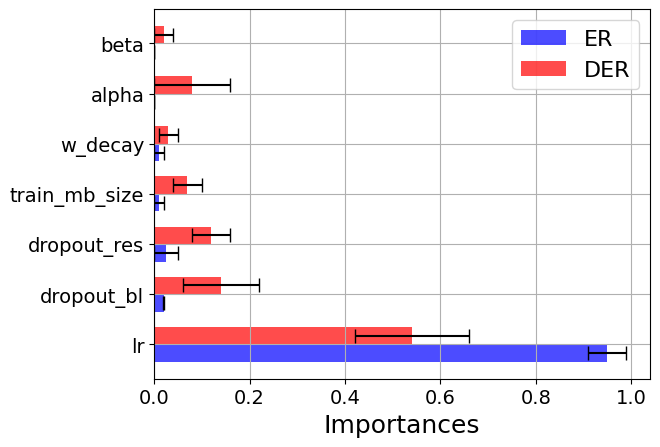}
  \end{subfigure}%

   \begin{subfigure}[b]{0.4\textwidth}
    \captionsetup{justification=centering}
    \caption{\label{fig:imp_er_online} ER and DER, Split-Tiny}
    \includegraphics[width=\textwidth]{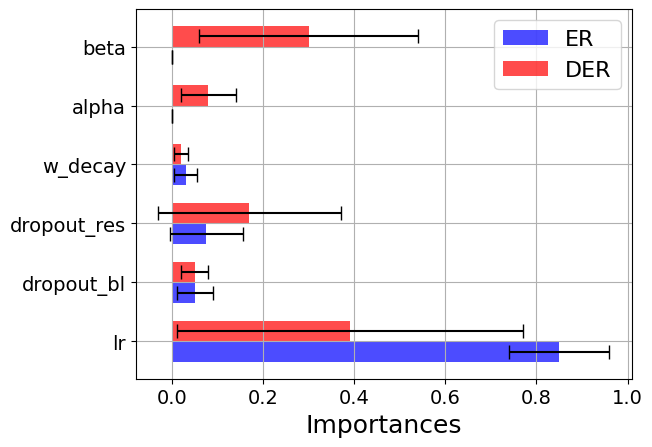}
  \end{subfigure}%
  \hspace{0.75cm}
  \begin{subfigure}[b]{0.4\textwidth}
    \captionsetup{justification=centering}
    \caption{\label{fig:imp_er_online} ER and DER, CORe50-NI}
    \includegraphics[width=\textwidth]{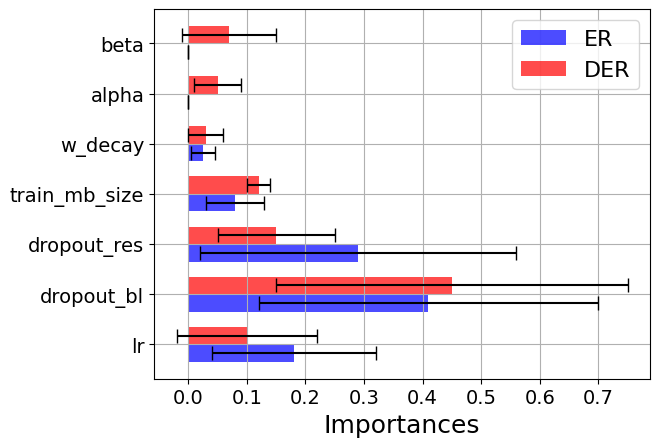}
  \end{subfigure}%

  \caption{Importance of hyperparameters averaged over the four continual learning benchmarks. We report the results for ER and DER continual strategies. The performance variability, estimated by fANOVA, is largely caused by a few hyperparameters that help define a subspace to which we can restrict configuration search space.}
  \label{fig:imp_er}
\end{figure*}
%%% IMP AVG  %%%

Currently, the family of methods with the best performance are memory-based methods. 
Due to this, we perform experiments using the ER and DER strategies for all the continual benchmarks presented in Subsection \ref{sec:benchmarks-metric}.
The performance variability primarily hinges on one or two hyperparameters, as we expected.
As shown in Figure \ref{fig:imp_er}, the learning rate plays a crucial role in the overall sequence accuracy performance in most incremental settings. 
In all the scenarios, the least important hyperparameter for final accuracy appears to be weight decay and changing it does not lead to significantly better results. 
Despite DER having a larger hyperparameter space than ER, similarities in hyperparameter importance are observed, as depicted in Figure \ref{fig:imp_er}. 
Contrary to expectations, specific continual hyperparameters in DER show a modest impact on performance, with beta significantly affecting only the Tiny benchmark.
Moreover, the experiments do confirm that by using the fANOVA approach we can have quantitative information on the most important hyperparameters that capture the potential for performance improvement over the sequence.

We already know that the choice of learning rate is a critical aspect that significantly influences the model's ability to adapt to new tasks without forgetting previously learned information.
%Indeed, the learning rate determines the size of steps taken during the optimization process, affecting the speed and stability of the model's updates. 
In Tiny and Rot-MNIST benchmarks, with more number of tasks, we note that the learning rate decreases over the sequence.
Furthermore, in Rot-MNIST, we note that the learning rate is more stable at the end of the sequence.
The intuition of the latter results is that the learning rate can balance the learning and forgetting of the sequence. 
In continual learning scenarios, decreasing the learning rate can mitigate catastrophic forgetting on future tasks, due to the reduction in the modification of weights, causing an equilibrium between plasticity and stability.
%Increasing the learning rate for new tasks can expedite learning and enhance adaptability while decreasing it for previously learned tasks helps mitigate forgetting, striking a balance between plasticity and stability in continual learning scenarios.
In CORe50 benchmark, a prevalent observation is the inclination towards increasing learning rate values. We believe that this tendency is motivated by the need for the model to swiftly adapt to new tasks without compromising its performance on previously learned ones. 
%A higher learning rate facilitates rapid parameter adjustments, aiding in the efficient incorporation of task-specific information.  
Choosing a DIL benchmark with no high number of tasks, the tendency to learn more in the new tasks seems reasonable.

This empirical trend highlights the importance of dynamic hyperparameter tuning, emphasizing the balance between adaptability and stability in the face of evolving data distributions.

\subsubsection{Hyperparameter Importance on Per Task Performance}

%%% Importance HPs %%%
% Online
\begin{figure*}[ht!]
  \centering

  \begin{subfigure}[b]{0.34\textwidth}
    \captionsetup{justification=centering}
    \caption{\label{fig:cifar_imp_er} ER, Split-CIFAR10}
    \includegraphics[width=\textwidth]{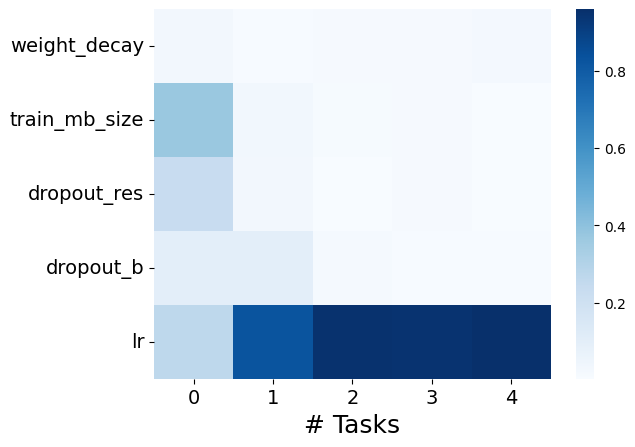}
  \end{subfigure}%
  \hspace{0.5cm}
  \begin{subfigure}[b]{0.49\textwidth}
    \captionsetup{justification=centering}
    \caption{\label{fig:cifar_imp_er_rot} ER, Rot-MNIST}
    \includegraphics[width=\textwidth]{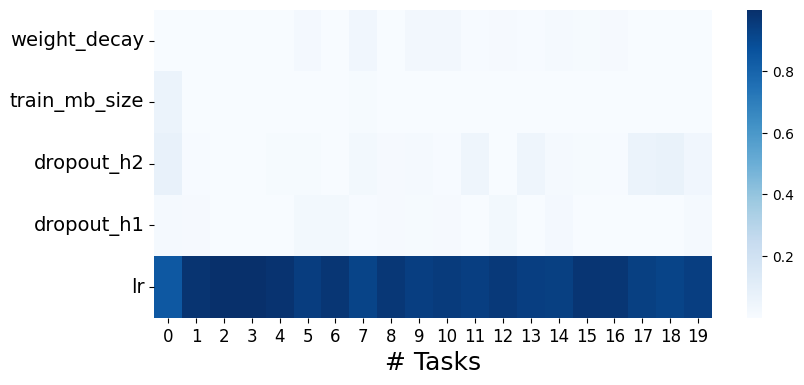}
  \end{subfigure}%
   \hfill

   \begin{subfigure}[b]{0.34\textwidth}
    \captionsetup{justification=centering}
    \caption{\label{fig:cifar_imp_er} ER+SI, Split-CIFAR10}
    \includegraphics[width=\textwidth]{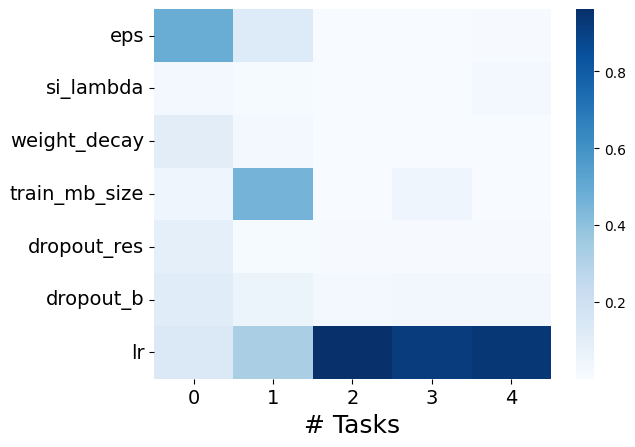}
  \end{subfigure}%
  \hspace{0.5cm}
  \begin{subfigure}[b]{0.49\textwidth}
    \captionsetup{justification=centering}
    \caption{\label{fig:cifar_imp_er_rot} ER+SI, Rot-MNIST}
    \includegraphics[width=\textwidth]{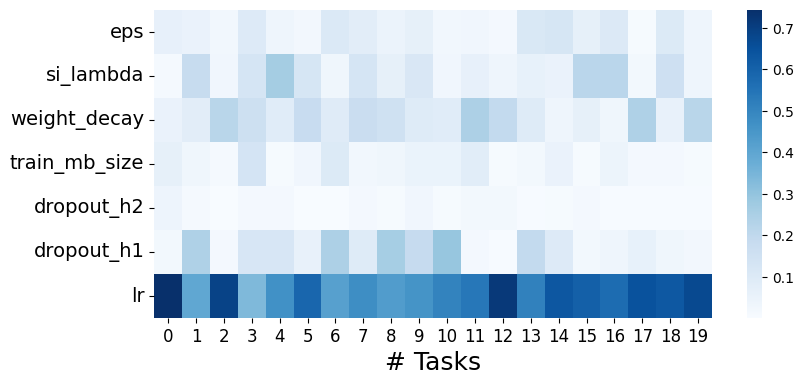}
  \end{subfigure}%
   \hfill

   \begin{subfigure}[b]{0.34\textwidth}
    \captionsetup{justification=centering}
    \caption{\label{fig:cifar_imp_er} ER+LwF, Split-CIFAR10}
    \includegraphics[width=\textwidth]{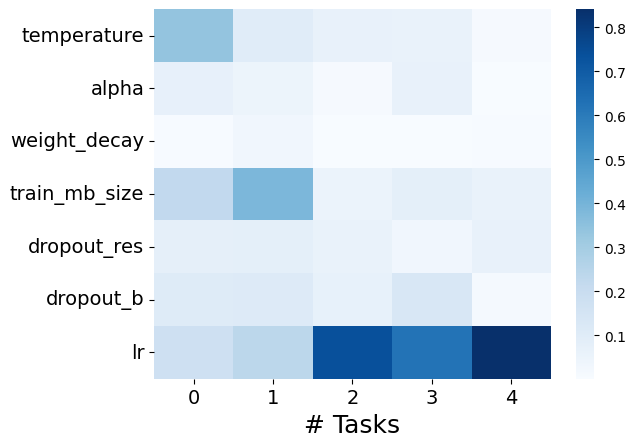}
  \end{subfigure}%
  \hspace{0.5cm}
  \begin{subfigure}[b]{0.49\textwidth}
    \captionsetup{justification=centering}
    \caption{\label{fig:cifar_imp_er_rot} ER+LwF, Rot-MNIST}
    \includegraphics[width=\textwidth]{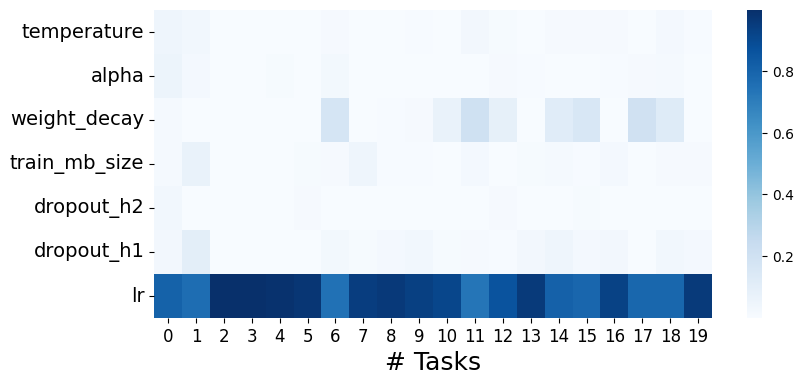}
  \end{subfigure}%

  \caption{Importance values for incremental per-task learning using ER and combined ER with SI and LwF; AdamW optimizer and TPE as Hyperparameter optimization (hpo). The results reported on the Online benchmarks show that a small set of hyperparameters is responsible for the most variation in performance. The learning rate plays an important role in online incremental scenarios.}
  \label{fig:online_imp_seq}
\end{figure*}

% Batch
\begin{figure*}[h]
  \centering

    \begin{subfigure}[b]{0.41\textwidth}
    \captionsetup{justification=centering}
    \caption{\label{fig:cifar_imp_er} ER, Split-Tiny}
    \includegraphics[width=\textwidth]{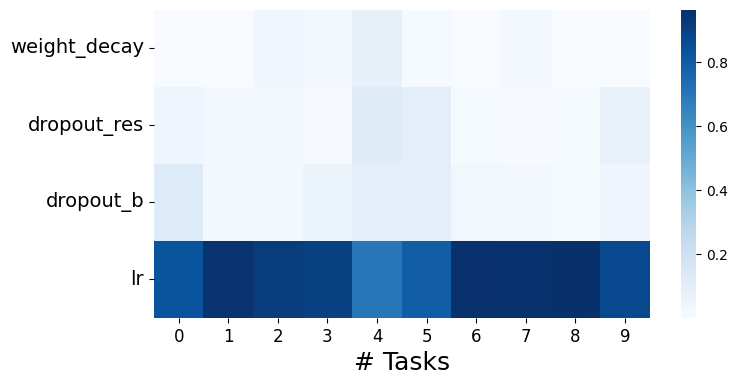}
  \end{subfigure}
  \hspace{0.5cm}
  \begin{subfigure}[b]{0.41\textwidth}
    \captionsetup{justification=centering}
    \caption{\label{fig:cifar_imp_er_rot} ER, CORe50-NI}
    \includegraphics[width=\textwidth]{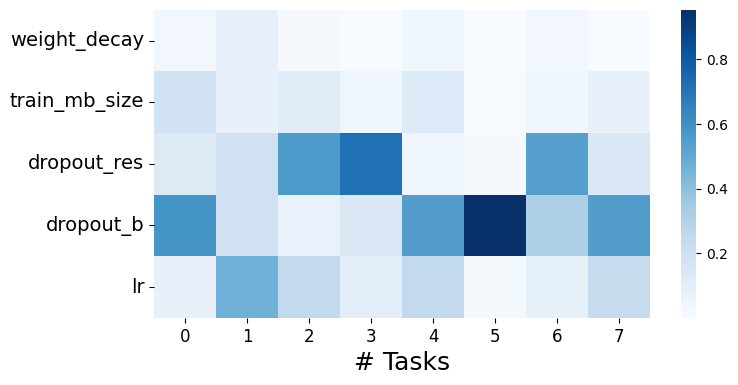} 
  \end{subfigure}
   \hfill

   \begin{subfigure}[b]{0.41\textwidth}
    \captionsetup{justification=centering}
    \caption{\label{fig:cifar_imp_er} ER+SI, Split-Tiny}
    \includegraphics[width=\textwidth]{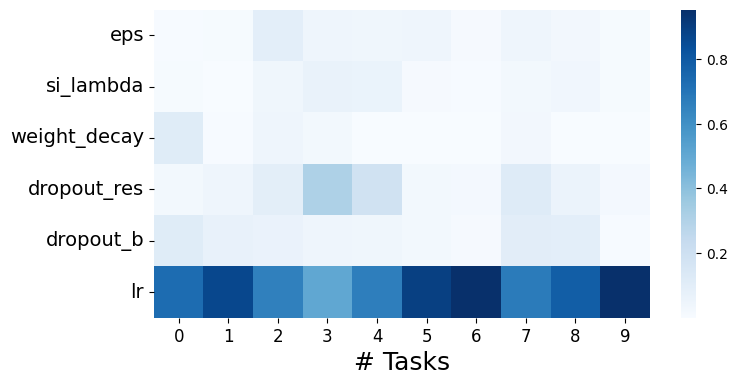}
  \end{subfigure}
  \hspace{0.5cm}
  \begin{subfigure}[b]{0.41\textwidth}
    \captionsetup{justification=centering}
    \caption{\label{fig:cifar_imp_er_rot} ER+SI, CORe50-NI}
    \includegraphics[width=\textwidth]{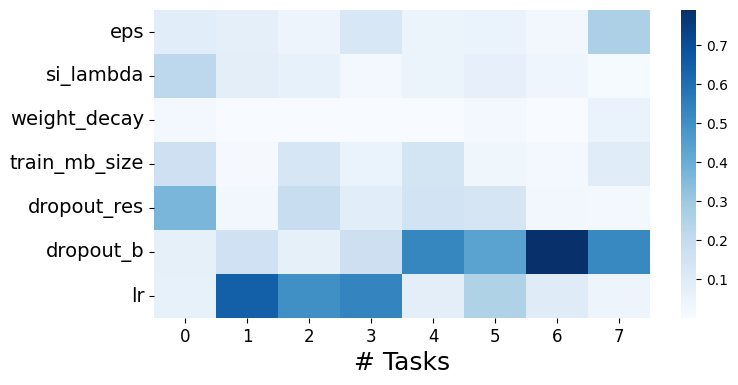} 
  \end{subfigure}
   \hfill

   \begin{subfigure}[b]{0.41\textwidth}
    \captionsetup{justification=centering}
    \caption{\label{fig:cifar_imp_er} ER+LwF, Split-Tiny}
    \includegraphics[width=\textwidth]{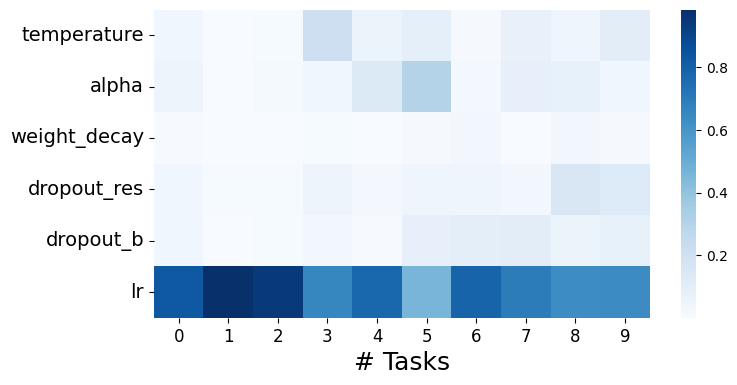}
  \end{subfigure}
  \hspace{0.5cm}
  \begin{subfigure}[b]{0.41\textwidth}
    \captionsetup{justification=centering}
    \caption{\label{fig:cifar_imp_er_rot} ER+LwF, CORe50-NI}
    \includegraphics[width=\textwidth]{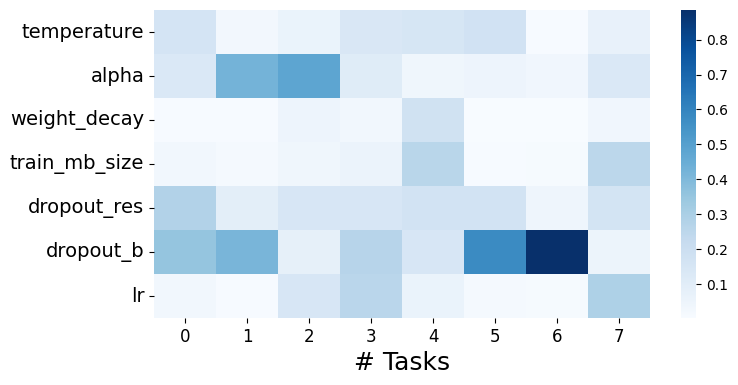}
  \end{subfigure}%

  \caption{Importance values for incremental per-task learning using ER and combined ER with SI and LwF; AdamW optimizer and TPE as Hyperparameter optimization (hpo). The results reported on the Batch benchmarks show that a small set of hyperparameters is responsible for the most variation in performance. }
  \label{fig:imp_batch_seq}
\end{figure*}
%%% END of Importance HPs %%%

\subsubsection*{Online Incremental Scenarios}

In this analysis, we provide an in-depth study of hyperparameter importance per task on ER, ER combined with SI and LwF to our selected online benchmarks CIFAR10 and Rot-MNIST.
We are interested in how the importance of a selected configuration of hyperparameters differs across different one-epoch learning tasks.

Following previous results, only a small set of hyperparameters is responsible for the most variation in performance, as shown in Figure \ref{fig:online_imp_seq}.
In this set of experiments, the most important parameter is the same which is the learning rate.
The fact that the proposed methods recovered this known most important hyperparameter also verifies that the proposed methodology works as expected.
However, the set of most important parameters is not the same over the sequences which does not validate our pre-experimental idea.
These results pose that having a method to change adaptively the hyperparameters during the sequence is a crucial aspect to take into account and fixing all the hyperparameters in all the sequences could hurt performance. 
Moreover, not all hyperparameters change at the same rate, that why it is important to use an incremental HPO method that knows how to mitigate the problem of testing all of them.
%We note that while it is well-known that learning rate is an important hyperparameter for online settings, to the best of our knowledge, this is the first study that provides systematic empirical evidence for their importance on continual benchmarks and strategies. The fact that the proposed methods recovered this known most important hyperparameter also verifies that the proposed methodology works as expected.

\subsubsection*{Batch Incremental Scenarios}

The conducted experiments for batch incremental scenarios (Tiny, CORe50) are shown in Figure \ref{fig:imp_batch_seq} with the same continual strategies chosen for online cases. 
%Given a classical or hybrid ER strategy with its associated set of hyperparameters, the results seem to demonstrate also in batch cases that performance variability is often largely caused by a few hyperparameters that define a subspace to which we can restrict configuration. 
The results seem to demonstrate also in batch cases that performance variability is often largely caused by a few hyperparameters that define a subspace to which we can restrict configuration.

In Tiny, the learning rate is the most important and plays a crucial role in all the continual strategies under study. 
We believe that in this benchmark, with more classes per task, it is reasonable that the learning rate decreases over the sequence to reduce catastrophic forgetting and have a high impact on performance as the fANOVA has quantified.
In CORe50, the dropout hyperparameters play a crucial role in most cases followed by the learning rate.
We believe that how the importance of hyperparameters in Tiny and CORe50 benchmarks change could be attributed to differences in scenarios (CIL and DIL), dataset sizes, and task complexities. The learning rate seems to dominate in scenarios where data or epochs are scarce, while dropout hyperparameters become crucial in handling the intricacies of diverse tasks in larger datasets preventing overfitting by randomly deactivating some neurons during training, enhancing the model's ability to generalize across various tasks.
To conclude, our findings highlight that, akin to an online setting, the quantitative importance evaluation effectively automates the selection of a hyperparameter subspace for each update step.

%%%%%
\subsection{Performance-Efficiency Analyses}

\begin{table*}[ht] % Utilizzo di table* per una tabella su due colonne
  \centering

  \begin{tabular}{cc|cc|cc}
    \hline
     & & \multicolumn{2}{c|}{Online} & \multicolumn{2}{c}{Batch} \\
    \hline
    Strategy & Method & \textbf{Split-Cifar10} & \textbf{Rotated-MNIST} &  \textbf{S-Tiny-ImageNet} & \textbf{CORe50-NI}  \\
    \hline
    \multirow{3}{*}{ER} & Fixed & $ 43.66 \pm 4.6 $ & $ 80.92 \pm 4.9 $ & $4.47 \pm  0.3$ & $ 46.10 \pm  9.4 $   \\
     & HPO & $53.67 \pm 0.3 $ & $ 84.55 \pm 0.6 $ & $24.40 \pm 1.0$ &  $ \mathbf{53.93 \pm  0.3} $   \\
    \rowcolor{myrowcolor}
     & Adaptive-HPO (our) & $ \mathbf{55.91 \pm 1.4} $ & $ \mathbf{84.92 \pm 2.6} $ & $ \mathbf{24.54 \pm  0.6} $ & $ 53.92 \pm  1.8 $   \\
    \hline
    \multirow{3}{*}{ER+SI} & Fixed & $ 43.34 \pm 5.4 $ & $ 82.63 \pm 2.6 $ &  $4.01 \pm 0.1$ & $ 36.43 \pm 4.3 $ \\
    & HPO &  $ \mathbf{57.32 \pm 0.4} $  & $ 84.44 \pm 0.7 $ & $ \mathbf{25.36 \pm 0.3} $ & $ \mathbf{48.48 \pm 0.9} $ \\
    \rowcolor{myrowcolor}
     & Adaptive-HPO (our) & $ 56.71 \pm 2.6 $ & $ \mathbf{85.14 \pm 3.0} $ & $ 24.72 \pm  0.8 $ & $ 45.74 \pm 4.7 $   \\
    \hline
    \multirow{3}{*}{ER+Lwf} & Fixed & $ 44.50 \pm 5.5 $ & $ 82.74 \pm 3.4 $ & $4.45 \pm 0.1$ & $ 36.08 \pm 4.4 $  \\
     & HPO & $ 55.03 \pm 0.3 $   & $ \mathbf{88.25 \pm 0.9} $ & $ \mathbf{26.00 \pm 1.4} $ & $ \mathbf{68.08 \pm 0.6} $   \\
    \rowcolor{myrowcolor}
     & Adaptive-HPO (our) & $ \mathbf{55.98 \pm 2.5} $  & $ 85.37 \pm 3.0 $ & $ 25.34 \pm  0.7 $ & $ 59.98 \pm 1.5 $  \\
    \hline
    \multirow{3}{*}{DER} & Fixed & $ 42.61 \pm 6.0 $ & $ 81.2 \pm 5.3 $ & $ 20.04 \pm 4.6 $ & $ 31.97 \pm 9.7 $  \\
     & HPO & $ \mathbf{56.2 \pm 0.8} $ & $ 86.42 \pm 0.3 $ & $ \mathbf{26.48 \pm 0.4} $ & $ 53.02 \pm 0.7 $  \\
    \rowcolor{myrowcolor}
     & Adaptive-HPO (our) & $ 55.71 \pm 1.0 $  & $ \mathbf{87.13 \pm 2.1} $ & $ 25.75 \pm 0.3 $ & $ \mathbf{53.19 \pm 3.8} $  \\
    \hline
    \multirow{3}{*}{GDumb} & Fixed & $ 11.96 \pm 2.6 $ & - & $ 3.50 \pm 5.0 $ & -  \\
     & HPO & $ \mathbf{22.44 \pm 0.5} $ & - & $ 14.19 \pm 0.9 $ & -  \\
    \rowcolor{myrowcolor}
    & Adaptive-HPO (our) & $ 20.46 \pm 1.6 $ & - &  $ \mathbf{14.60 \pm 0.5}  $ & -  \\
    \hline
  \end{tabular}
  
  \caption{Final stream accuracy results for all four CL benchmarks averaged across 3 runs. ‘-’ indicates experiments we were unable to run or unimportant for our purposes (e.g. Gbumb in DIL benchmarks).  Fixed: random choice of hyperparameters in the first task and keep fixed in the remaining; HPO: use hpo for each task;  Adaptive-HPO: our proposed solution.  For Online the buffer memory size is 500 while for batch 5120. Optimizing adaptively outperforms fixed tuning by a large margin, especially in the most difficult benchmark Tiny. In all benchmarks, our solution achieves a small gap in the worst case concerning the standard time-consuming greedy HPO approach.}
  \label{tab:table_finalSA}
\end{table*}

\begin{figure}[h]
  \centering
  \includegraphics[width=0.45\textwidth]{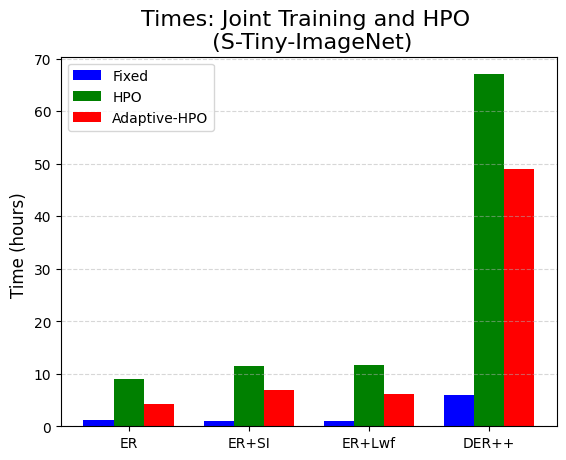}
  \caption{Results for executing training and optimization jointly in terms of the time demand on Tiny (over all the sequence); [$\downarrow$] lower is better}
  \label{fig:eff_tiny}
\end{figure}

\begin{figure}[h]
  \centering
  \includegraphics[width=0.45\textwidth]{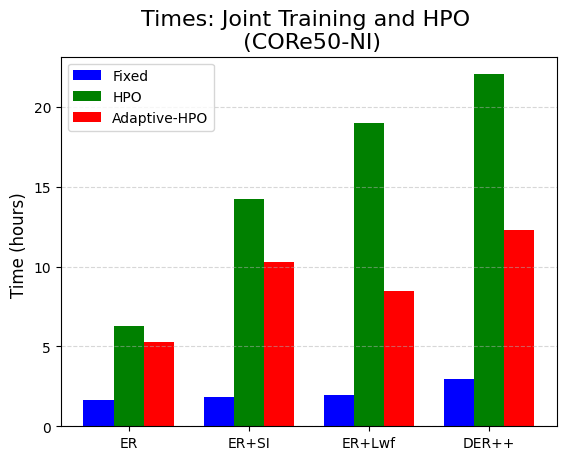}
  \caption{Results for executing training and optimization jointly in terms of the time demand on CORe50; [$\downarrow$] lower is better}
  \label{fig:eff_core}
\end{figure}

This section delves into the discussion of experimental results achieved for \textbf{(RQ2)}, with further discussion on our early hypotheses provided in Subsection \ref{subsec:rq2}.
First, we study the final stream accuracy of the Fixed and HPO then we investigate Adaptive-HPO (our solution) to compare with the two baselines. The results are presented in Table \ref{tab:table_finalSA}. The results of the efficiency analysis are in Figure \ref{fig:eff_tiny} and Figure \ref{fig:eff_core}.

\subsubsection{Performance Analyses}
We can observe that both Adaptive HPO and HPO outperform Fixed baseline in terms of performance for all four benchmarks.
In Tiny our method outperforms the Fixed baseline by $+20 \% $ for all the continual strategies under study.
In CORe50, the Fixed baseline presents a high variance in performance over different seeds showing a high sensitivity to the hyperparameters selection and that a wrong selection and keeping fixed them over the sequence have a high impact on performance.
The results maintain equal increasing the number of random seeds.
We can argue that doing per-task HPO increase the overall accuracy of methods, as we have expected.

Interestingly, Adaptive-HPO appears to surpass HPO in some benchmarks or continual strategies. For instance, in Rot-MNIST experiments, we achieve the highest accuracy score in ER, ER+SI and DER. Superior performance results appear in most cases also for ER strategy.
It is pertinent to consider that HPO operates as a greedy algorithm, seeking the best hyperparameters for the current task. 
The optimal hyperparameters identified through HPO for a specific task may not necessarily be the most optimal for the entire sequence of tasks. 
This suggests that the ideal set of hyperparameters can dynamically shift between tasks, deviating from what might be deemed optimal for the entire sequence.
By employing a quantitative assessment of hyperparameter importance through fANOVA, we know that not all hyperparameters change at the same variance. 
This underscores the significance of avoiding greedy HPO methods to address the challenge of selecting optimal configurations incrementally within expansive search spaces.

\subsubsection{Efficiency Analyses}
When facing up with an incremental stream, one often cares about reducing the overall processing training time otherwise, training would not keep up with the rate at which data are made available to the stream. We extend this analysis to perform training and HPO jointly given a sequence of learning tasks.
In particular, our Adaptive-HPO is compared with the baselines with empirical results in our most time-consuming benchmarks, Tiny in Figure \ref{fig:eff_tiny} and CORe50 in Figure \ref{fig:eff_core}.
We draw two main remarks. First, we note that DER processes stream data slowly making it hardly viable in practical batch HPO scenarios.
Second, in Tiny the time complexity for our Adaptive-HPO is constant over all the tasks and reduces the baseline HPO time demand by 53\% for ER, 39\% for ER+SI, 47\% for ER+LwF and 27\% for DER.
In CORe50 the time complexity for our Adaptive-HPO is reduced by the baseline HPO time demand by 16\% for ER, 28\% for ER+SI, 55\% for ER+LwF and 44\% for DER.

\subsubsection{Final Remarks on Performance-Efficiency Analyses}
In our experiments, by executing a sophisticated HPO for each model update, we note that the selection of a combination of continual strategies positively affects performance accuracy and it is possible automatically.
Moreover, the results shown in Table \ref{tab:table_finalSA}  empirically demonstrate that the hyperparameters determined as the most important ones indeed are the most important ones to optimize in sequence task learning by further benefit on time demand. Performing hyperparameter optimization in all the sequences improves the final accuracy in diverse scenarios, including Online, Batch, CIL, and DIL.

%%%%%
\subsection{Robustness Analyses}

\begin{figure*}[h!]
  \centering
  
  \begin{subfigure}[b]{0.425\textwidth}
    \captionsetup{justification=centering}
    \caption{\label{fig:imp_erlwf_tiny} ER+SI, Split-CIFAR10}
    \includegraphics[width=\textwidth]{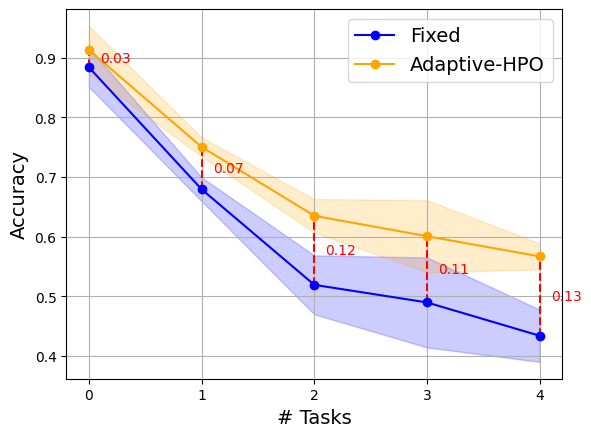}
  \end{subfigure}% 
  \hspace{0.5cm}
  \begin{subfigure}[b]{0.425\textwidth}
    \captionsetup{justification=centering}
    \caption{\label{fig:imp_er_core} ER+SI, Split-Tiny}
    \includegraphics[width=\textwidth]{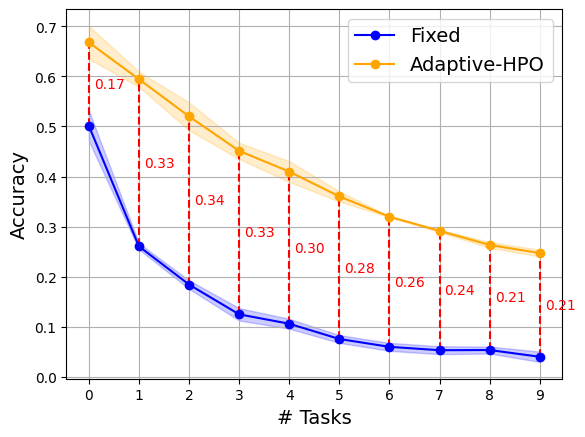}
  \end{subfigure}

  \begin{subfigure}[b]{0.425\textwidth}
    \captionsetup{justification=centering}
    \caption{\label{fig:imp_erlwf_tiny} ER+LwF, Split-CIFAR10}
    \includegraphics[width=\textwidth]{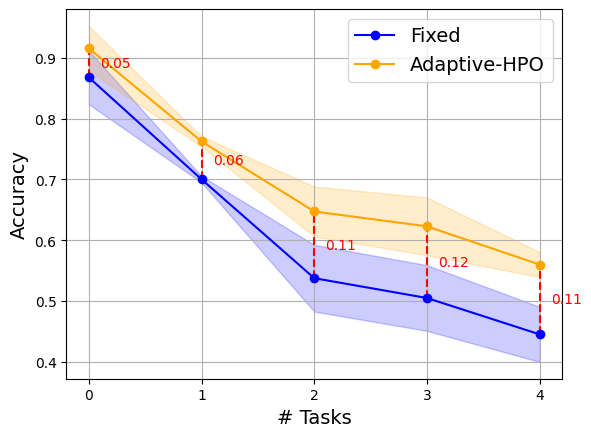}
  \end{subfigure}% 
  \hspace{0.5cm}
  \begin{subfigure}[b]{0.425\textwidth}
    \captionsetup{justification=centering}
    \caption{\label{fig:imp_er_core} ER+LwF, Split-Tiny}
    \includegraphics[width=\textwidth]{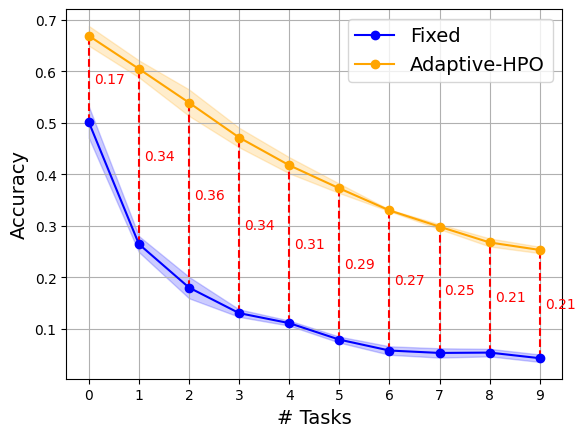}
  \end{subfigure}

  \caption{Results for the model robustness analysis comparing Fixed baseline with our Adaptive-HPO solution on CIFAR-10 and TinyImageNet for ER combined with SI and LwF. Optimizing adaptively outperforms fixed selection by a large margin, across all incremental learning steps.}
  \label{fig:rob_ersilwf}
\end{figure*}

In our last analysis, we delve into the discussion of experimental results achieved for \textbf{(RQ3)}, with further discussion on our early hypotheses provided in Subsection \ref{subsec:rq3}.
We compare the Fixed baseline with our Adaptive-HPO versions on ER+SI and ER+LwF continual strategies. 
Results are presented in Figure \ref{fig:rob_ersilwf}.

As can be seen, regardless of the strategies, adaptivity leads to a significant increase in per-task accuracy, leading up to $13\%$ for ER+SI and $12\%$ for ER+LwF on CIFAR10,  while  $34\%$ for ER+SI and $36\%$ for ER+LwF on Tiny. 
This led us to observe that tuning hyperparameters when we update the model with new data improves the per-task accuracy performance. 
Furthermore, this confirms our hypothesis that incremental learners can benefit greatly from learning and tuning hyperparameters jointly to the current task while keeping fixed parameters leads to worse performance.
The results also confirm that performing HPO for each model update improves the robustness concerning the sequence order of tasks and enhances performance trends across the stream.

%%%%%%%%%%%%%%%%%%%%%%%%%%%%%%%%%%%%%%%%%%%%%%%%%%%%%%%%%%%%%%%%%%%%%%%%%%%%%%%%%%%%%%%%%%%%%%%%%%%%%%%%%%%%%%%%%%%%%%%%%%%%%%%
%  
%%%%%%%%%%%%%%%%%%%%%%%%%%%%%%%%%%%%%%%%%%%%%%%%%%%%%%%%%%%%%%%%%%%%%%%%%%%%%%%%%%%%%%%%%%%%%%%%%%%%%%%%%%%%%%%%%%%%%%%%%%%%%%%
\section{Conclusion}
In this paper, we study the critical challenge of hyperparameter selection in continual learning scenarios, especially within non-stationary environments. 
Employing an innovative approach grounded in functional analysis of variance (fANOVA), we identify automatically key hyperparameters that have an impact on performance. 
We demonstrate empirically that this approach, agnostic to continual scenarios and strategies, allows the speed up of hyperparameters optimization adaptively across tasks.

Our experiments, conducted on CIFAR10, Rot-MNIST, Tiny, and CORe50, have yielded three findings. 
First, dynamically adapting hyperparameters during the sequence is crucial, underscoring the importance of an incremental method for hyperparameter optimization brings significant improvements in accuracy and transfer, regardless of the continual strategy used. 
Second, the automatic selection of hyperparameters has highlighted substantial improvements in terms of average accuracy and computational cost, striking a balance between performance efficiency.
Thirdly, tuning hyperparameters during model updates has proven instrumental in enhancing per-task accuracy performance. Notably, our strategy exhibits robustness even in the face of varying sequential task orders.

These results suggest that incorporating the hyperparameter update rule and automatic selection per task significantly bolsters the adaptation process on streaming data. We hope our work will inspire future research for further contributing to the advancement of continual learning methodologies and automating the customization of parameters in incremental learning. This progress aims at fostering the development of more efficient, robust, and adaptable models for real-world applications.

%%%%%%%%%%%%%%%%%%%%%%%%%%%%%%%%%%%%%%%%%%%%%%%%%%%%%%%%%%%%%%%%%%%%%%%%%%%%%%%%%%%%%%%%%%%%%%%%%%%%%%%%%%%%%%%%%%%%%%%%%%%%%%%
% REFERENCES 
%%%%%%%%%%%%%%%%%%%%%%%%%%%%%%%%%%%%%%%%%%%%%%%%%%%%%%%%%%%%%%%%%%%%%%%%%%%%%%%%%%%%%%%%%%%%%%%%%%%%%%%%%%%%%%%%%%%%%%%%%%%%%%%
\bibliographystyle{plain}
\bibliography{reference}

\vfill

\end{document}